\documentclass[10pt,twocolumn,letterpaper]{article}

\usepackage[pagenumbers]{iccv} 

\definecolor{iccvblue}{rgb}{0.21,0.49,0.74}
\usepackage[pagebackref,breaklinks,colorlinks,allcolors=iccvblue]{hyperref}

\usepackage{dsfont}
\usepackage{bm}
\usepackage{multirow}
\usepackage{adjustbox}
\usepackage{amsmath}
\usepackage{float}
\usepackage{caption}
\usepackage{amssymb}
\usepackage{subcaption}
\usepackage{verbatim}
\usepackage{multicol}
\usepackage{booktabs}
\usepackage{ulem}
\usepackage{comment}
\usepackage{soul,color} 
\usepackage{kotex} 
\usepackage{pifont}
\usepackage{multirow}
\usepackage{pifont}
\newcommand{\cmark}{\ding{51}}
\newcommand{\xmark}{\ding{55}}
\usepackage[linesnumbered,ruled]{algorithm2e}

\def\paperID{13470} 
\def\confName{ICCV}
\def\confYear{2025}

\title{TAG: A Simple Yet Effective Temporal-Aware Approach for Zero-Shot Video Temporal Grounding}

\author{Jin-Seop Lee\thanks{Equal contribution}\qquad SungJoon Lee\footnotemark[1]\qquad Jaehan Ahn\qquad YunSeok Choi\qquad Jee-Hyong Lee\thanks{Corresponding author}\\
Sungkyunkwan University, Suwon, South Korea\\
{\tt\small \{wlstjq0602, sjoon8379, ajh508, ys.choi, john\}@skku.edu} 
}

\begin{document}
\maketitle
\begin{abstract}
Video Temporal Grounding (VTG) aims to extract relevant video segments based on a given natural language query. 
Recently, zero-shot VTG methods have gained attention by leveraging pretrained vision-language models (VLMs) to localize target moments without additional training. 
However, existing approaches suffer from semantic fragmentation, where temporally continuous frames sharing the same semantics are split across multiple segments. When segments are fragmented, it becomes difficult to predict an accurate target moment that aligns with the text query. Also, they rely on skewed similarity distributions for localization, making it difficult to select the optimal segment. Furthermore, they heavily depend on the use of LLMs which require expensive inferences.
To address these limitations, we propose a \textit{TAG}, a simple yet effective Temporal-Aware approach for zero-shot video temporal Grounding, which incorporates temporal pooling, temporal coherence clustering, and similarity adjustment.
Our proposed method effectively captures the temporal context of videos and addresses distorted similarity distributions without training. Our approach achieves state-of-the-art results on Charades-STA and ActivityNet Captions benchmark datasets without rely on LLMs. 
Our code is available at \href{https://github.com/Nuetee/TAG}{\texttt{github.com/Nuetee/TAG}}.

\end{abstract}
\section{Introduction}
\label{sec:intro}

Platforms like YouTube have become essential digital resources, offering vast video content for users to explore. However, manually locating specific segments within lengthy videos remains a time-consuming and labor-intensive task. Video Temporal Grounding (VTG) aims to automatically extract relevant segments from videos given a natural language query. 

Previous VTG approaches have been proposed to train models using pairs of natural language queries and their corresponding video moments~\cite{zhang2020learning, huang2022video, liu2022memory, jang2023knowing, huang2021cross, zheng2022weakly_aaai, zheng2022weakly_cvpr, huang2023weakly}. However, 
well-annotated paired datasets is expensive 
to produce, and models trained with such datasets demonstrate poor generalization performance as the datasets have a limited set of videos and texts.
To address these issues, there has been growing interest in the field of zero-shot video temporal grounding, which leverages pretrained vision-language models (VLMs) to localize the target moment.

Since VLMs are pretrained on large datasets of image and text pairs, zero-shot video temporal grounding (ZSVTG) methods based on VLMs demonstrate strong generalization performance without additional fine-tuning~\cite{wacv, vtggpt, tfvtg}. To localize the target moment, they first generate candidate segments either by clustering video frame features~\cite{wacv} or by heuristic approaches (e.g., uniformly dividing the video into pre-defined intervals)~\cite{vtggpt,tfvtg}.  Then, the most relevant segment is selected based on the similarity between the segments and the given text query.

\begin{figure*}[t]
\flushleft
\includegraphics[width=0.9\linewidth]{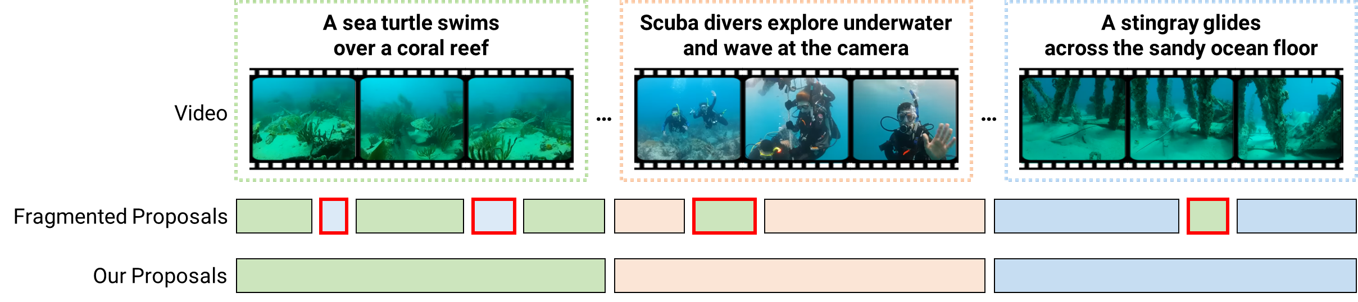}
\vspace{0.1cm}
\caption{Motivation of our proposed method. Existing approaches tend to generate fragmented proposals, where semantically continuous video segments are divided into multiple disjoint proposals (red boxes). This is due to a lack of consideration for temporal context and temporal coherence during proposal generation.}
\label{fig1}
\vspace{-0.2cm}
\end{figure*}

While these approaches~\cite{wacv,vtggpt,tfvtg} can localize segments that roughly correspond to the query, they often suffer from \textit{semantic fragmentation}, where temporally continuous frames sharing the same semantics are split across multiple segments. Although adjacent video frame features which belong to the same action or scene, transient noise (e.g., variations in camera angle or lighting) can cause inconsistencies in their visual representations, as shown in \cref{tsne}(a). As a result, they are assigned to different clusters, and their proposals are split regardless of the context, as illustrated in \cref{fig1}.
To mitigate semantic fragmentation, it is necessary to consider temporal context and enhance temporal coherence to accurately predict the target moment.

\vspace{0.2cm}

Another limitation is that most methods simply use alignment similarities without considering the similarity distribution when selecting the candidate segment most relevant to the query. When either a majority or a minority of frames in the video are close to the query, this can lead to skewed similarity distributions. Such skewed distributions may hinder the selection of the optimal segment, resulting in a decline in overall performance. To address this, adaptive similarity adjustment based on the similarity distribution is required.

We present TAG, a simple yet effective \underline{T}emporal-\underline{A}ware approach for zero-shot video temporal \underline{G}rounding.
To effectively capture the temporal context of images within videos, we propose temporal pooling and temporal coherence clustering. In temporal pooling, we incorporate temporal information into the extracted image features by aggregating features from adjacent images. Based on these temporally aggregated features, we generate contextual proposals by temporal coherence clustering. These methods help the model effectively capture the video context and generate boundary-aligned candidate proposals. Also, To prevent performance degradation by the distorted similarity distribution when selecting the most suitable proposal, we propose similarity adjustment. This transformation normalized similarities, where higher values are amplified, and lower values are dampened.

Notably, TAG outperforms existing approaches that heavily rely on large language models (LLMs), despite not using LLMs. This demonstrates that our method is not only simple and effective, but also cost-efficient. We validate our approach through experiments on the Charades-STA~\cite{gao2017tall} and ActivityNet Captions~\cite{krishna2017dense} datasets and various scenarios. We achieve state-of-the-art performance across all datasets and scenarios, with up to 2.65\% mIoU improvement on Charades-STA and 7.18\% on ActivityNet Captions in general settings.

\begin{figure*}[t]
\centering
\includegraphics[width=0.95\linewidth]{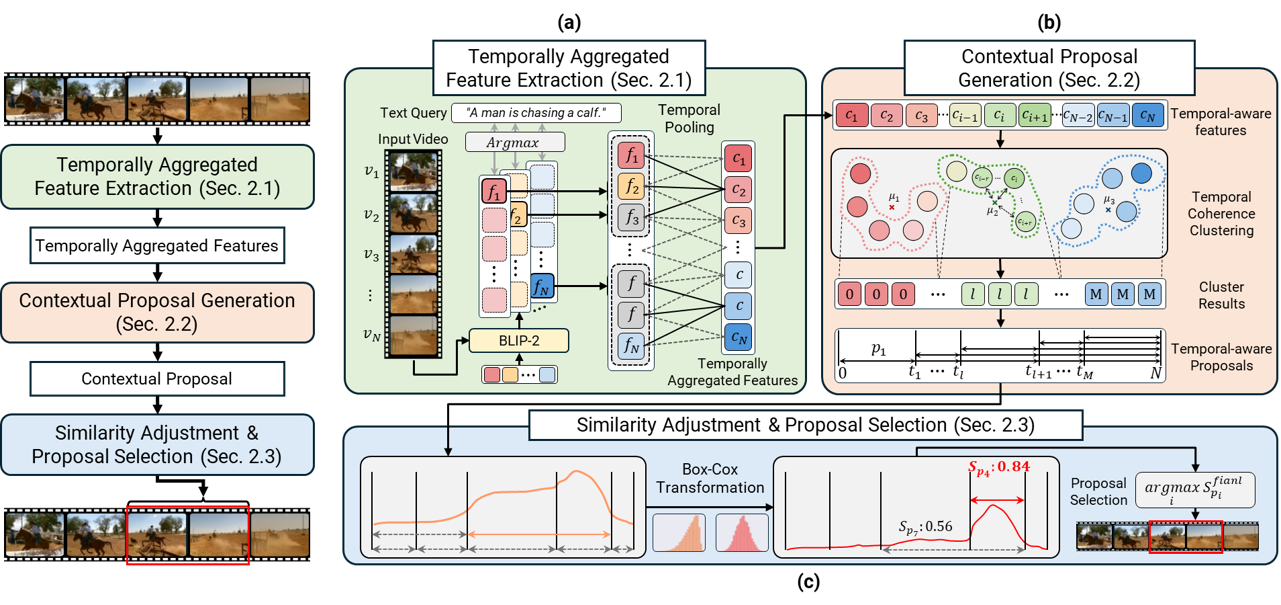}
\vspace{-0.4cm}
\caption{\textbf{Overview of our proposed method}. We first incorporate temporal dependencies of consecutive features using a temporal pooling (\cref{sec3_1}). Based on this, we extract temporally aggregated features. Then, we generate contextual proposals. To effectively generate segment proposals, we propose temporal coherence clustering (\cref{sec3_2}). Lastly, we adjustment similarities, and then, select the most relevant proposal (\cref{sec3_3}).}
\label{framework}
\end{figure*}
\section{Related Work}
\paragraph{Training-based Video Temporal Grounding.}

Training-based VTG methods can be categorized into fully supervised, weakly supervised, and unsupervised learning.
Fully supervised VTG methods~\cite{zhang2020learning, huang2022video, liu2022memory, jang2023knowing, li2024groundinggpt, huang2024vtimellm, wu2020tree, yuan2019semantic, li2022compositional, yang2023deco} train models using pairs of natural language queries and their corresponding video segments with precisely annotated start and end boundaries. 
However, labeling video segments requires specifying the exact start and end frames, making the large-scale collection of text-query and video-segment datasets both costly and labor-intensive.
To address this issue, weakly supervised and unsupervised approaches have been proposed recently. Weakly supervised VTG methods~\cite{huang2021cross, zheng2022weakly_aaai, zheng2022weakly_cvpr, huang2023weakly, duan2018weakly} utilize textual descriptions of videos without requiring start or end frame annotations for each segment. On the other hand, unsupervised VTG methods~\cite{gao2021learning, nam2021zero, wang2022prompt, kim2023language, zheng2023generating} train models without textual descriptions and precise segment boundaries. 
These methods generate pseudo-text queries and treat them as textual descriptions, similar to weakly supervised approaches.

These training-based VTG methods perform well on videos within the trained distribution but often struggle with unseen scenarios.
This limitation arises from the labor-intensive process of accurately labeling segments for a given natural language query. The absence of large and diverse datasets that cover various scenarios and contexts further weakens the model's ability to generalize. Additionally, these approaches often require high computational costs for training, making them less practical for real-world applications.
To address these limitations, zero-shot video temporal grounding methods have been proposed, leveraging the strong generalization capabilities of pre-trained VLMs without additional training costs.

\paragraph{Zero-shot Video Temporal Grounding.}
Most zero-shot video temporal grounding methods leverage VLMs, which are pretrained on large datasets consisting of image and natural language sentence pairs. These methods treat a video as a set of independent images and perform localization based on the alignment similarities between the text query feature and the image features. 
\citet{wacv} first introduced the ZSVTG task and proposed a bottom-up proposal generation strategy utilizing VLMs. 
\citet{vtggpt} introduced VTG-GPT to generate video frame captions using GPT and create proposals by combining the generated captions with the input text query. 
On the other hand, \citet{tfvtg} proposed TFVTG to leverage large language models (GPT-4 Turbo)~\cite{achiam2023gpt} to generate paraphrased versions of the given query or to decompose it into multiple sub-queries.

\section{Proposed Method}
We aim to the model effectively capture the video's temporal context information and enhance temporal coherence without any training.
Also, we address the distorted similarity distributions and the reliance on LLMs.
To achieve this goal, we introduce a TAG, simple yet effective \underline{T}emporal-\underline{A}ware approach for zero-shot video temporal \underline{G}rounding. In this section, we introduce temporal pooling (\cref{sec3_1}), temporal coherence clustering (\cref{sec3_2}), and similarity adjustment (\cref{sec3_3}).

\begin{figure}[t]
\centering
\includegraphics[width=1\linewidth]{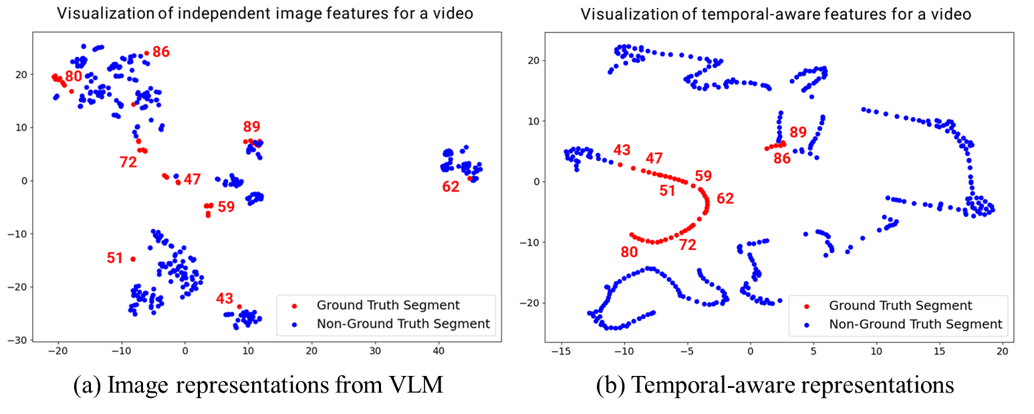}
\caption{T-SNE visualizations for a single video.  
Red and blue points indicate representations of the ground truth segment and other segments, respectively. The numbers represent frame indices.  
(a) Features extracted from VLM are scattered without regard to time order.  
(b) Temporal-aware features are arranged in temporal sequence.}
\label{tsne}
\vspace{-0.3cm}
\end{figure}

\subsection{Temporally Aggregated Feature Extraction}
\vspace{-0.0cm}
\label{sec3_1}

For a given video and text query, our first objective is to extract temporally aggregated features that incorporate temporal context information. We first extract single-frame image features using VLMs. As shown in \cref{framework}(a), we use pretrained BLIP-2 image encoder, text encoder, Q-Former, and learned queries~\cite{li2023blip}. 
Each image is extracted into multiple representations based on the number of learned queries, and among these, the image representation most similar to the text representation is selected.
Then, we could extract single-frame image features \( F = [f_1, f_2, \dots, f_N] \) for \(N\) frames. However, they still do not contain temporal context information. To inject the temporal-aware information into the features, we propose a temporal-aware pooling, which uses sliding window-based average pooling.

The temporal pooling is a non-parametric convolutional layer, with a large window size \( w \) and a stride of 1.
For frame \( i \), the aggregated feature \( c_i \) is computed by averaging features across consecutive frames, and it is as follows:
\begin{equation}
    c_i = \frac{1}{w} \sum_{j=i - (w-1)/2}^{i + (w-1)/2} f_j
\end{equation}
where \( w \) is the kernel window size, and \( f_j \) represents the feature of the \( j \)-th frame. 
With a large window size, it enables capturing a wider range of context information.
To maintain the original sequence length \( N \), we apply feature padding at the edges.

Through the temporal pooling, which aggregates temporal context information across consecutive image features, we can extract temporally aggregated features \( C = [c_1, \dots, c_N] \).
It is very simple yet effective at incorporating temporal context information into the model without any training.
As shown in \cref{tsne}(b), our features are not only located in the feature space according to the time order, but are also clearly separable based on temporal context.

\subsection{Contextual Proposal Generation}
\vspace{-0.0cm}
\label{sec3_2}

With the extracted temporally aggregated features \( C \), we aim to generate contextual proposals that align with context boundaries. To achieve this, we introduce temporal coherence clustering which enhances temporal coherence. Unlike conventional clustering methods~\cite{hartigan1979algorithm} that consider only individual frames, our method assigns clusters by also taking into account temporally neighboring frames. Our temporal coherence clustering objective is as follows:
\begin{equation}
\arg\min_{\mathcal{S}} \sum_{j=1}^{k} \sum_{c^{(i)} \in S_j} \sum_{\Delta = -(r-1)/2}^{(r-1)/2} \left\| c^{(i+\Delta)} - \mu_j \right\|^2,
\end{equation}
where \( c^{(i)} \) is the feature of the \( i \)-th frame, \( \mu_j \) is the centroid of the \( j \)-th cluster, and \( r \) indicates the temporal window size. When \( r = 0 \), the formulation is same as naive k-means clustering. This objective encourages temporally adjacent frames to be assigned to the same cluster when they exhibit similar features, thereby enhancing temporal coherence and identifying the points where the video's temporal context changes.

As shown in \cref{framework}(b), based on temporal coherence clustering, we obtain clustering results \( L = [l_1, \dots, l_N] \) for all aggregated features. 
Since we use temporal pooling to extract aggregated features, adjacent aggregated features are very similar. Also, temporal coherence clustering encourages adjacent frames to be assigned to the same cluster. So, if video content does not change, images will belong to the same cluster.
For all frames, we identify the points where the cluster labels change. These change points \( T \) are formulated as follows:
\begin{equation}
\begin{gathered}
T   = \{0, t_1, \cdots, t_M, N\}, \\
t_m = \min \{\, i \mid \{\, l_i \neq l_{i+1} \,\} \, \cap \, \{\, i > t_{m-1} \,\} \},\\
m   \in \{1, \cdots, M\}
\end{gathered}
\end{equation}
where \( l_i \) is the cluster label of frame \( i \), \( M \) is the total number of label changes.
Here, \( M \) is not a hyperparameter, which is determined from the clustering result.
Each \( t_m \) represents the \( m \)-th point where a cluster label change occurs, marking the boundary for segment proposals, and we add the start and end points of the frames to \( T \).

Based on these change points \( T \), we generate contextual proposals with diverse ranges. The generated contextual proposals are formulated as follows:
\begin{equation}
\begin{gathered}
P = \left\{ p_k \mid p_k = (t_i, t_j), \; t_i, t_j \in T, \; i < j \right\}, \\
k \in \left\{ 1, \cdots, \frac{(M+1)(M+2)}{2} \right\}
\end{gathered}
\end{equation}
where \( P \) is the set of contextual proposals, and \( t_i \) and \( t_j \) denote the start and end points of each segment, respectively. This results in \( (M+1)(M+2)/2 \) possible segments, providing diverse temporal ranges within the video. 

Our approach allows for generating more proposals for videos with frequent context changes and fewer proposals for videos with infrequent context changes. Additionally, since our candidate proposals are formed from all possible combinations of context boundaries, they have a diverse range of lengths.

\subsection{Similarity Adjustment \& Proposal Selection}
\label{sec3_3}

To localize the target moment, we need to choose the most relevant segment among the temporal-aware proposals. To evaluate the semantic relevance for segment proposals, we calculate the alignment similarities between the text query \( q \) and single-frame image features \( F \). At each \( i \)-th frame, the basic similarity is represented as \( f_i \cdot q \). 

However, the basic similarity-based proposal selection method does not consider the similarity distribution, which leads to skewed similarity distributions and hinders the selection of the optimal segment. To address this issue, we propose similarity adjustment.
To make the distribution closer to a normal distribution, we apply a Box-Cox transformation~\cite{box1964analysis}. The adjusted similarities are formulated as follows:
\begin{equation}
\text{a}_i = \frac{(f_i \cdot q)^\lambda - 1}{\lambda}, \enspace i \in \{1, 2, \cdots, N\}
\end{equation}
where \( \lambda \) is the parameter for the Box-Cox transformation, and \( f_m \cdot q \) represents the similarity between the feature \( f_m \) of frame \( m \) and the query \( q \). 
When similarities are normalized, it mitigates the skewness in the similarity distributions, as shown in \cref{framework}(c). This approach enables adaptive normalization based on the similarity distribution, which helps in selecting the optimal segment.
With these normalized similarities, we calculate the scores for all contextual proposals \(P\).
The proposal score we use is formulated as follows:
\begin{equation}
\begin{gathered}
S = \{ S_{p_i} \mid p_i = (t_i, t_j), \; p_i \in P \}, \\
S_{p_i} = \frac{1}{t_j - t_i} \sum_{m = t_i}^{t_j} a_m - \frac{1}{N - (t_j - t_i)} \sum_{n \notin [t_i, t_j)} a_n
\end{gathered}
\end{equation}

where \( S \) is the set of proposal scores \( S_{p_i} \) for each proposal \( p_i =
(t_i, t_j) \in P \), with \( t_i \) and \( t_j \) representing the start and end points of the proposal. These proposal scores represent the difference in relevance between the segment and the non-segment regions. 

Finally, based on the proposal scores, we rank all temporal-aware proposals, and select the proposal with the highest score as the final output. 
With our proposed method, the model effectively captures the temporal context of images within videos, and selects the most suitable proposal based on a non-distorted similarity distribution.
\begin{table*}[t]
\centering
\footnotesize
\renewcommand{\arraystretch}{0.95} 
\begin{tabular}{c|ccc|cccc|cccc}
\toprule
\multirow{2}{*}{Method} & \multirow{2}{*}{Setting} & \multirow{2}{*}{VLM} & \multirow{2}{*}{LLM} & \multicolumn{4}{c|}{Charades-STA} & \multicolumn{4}{c}{ActivityNet Captions} \\ \cline{5-12} \rule{0pt}{1.2EM}
& & & & R@0.3 & R@0.5 & R@0.7 & mIoU & R@0.3 & R@0.5 & R@0.7 & mIoU \\
\midrule
2D-TAN \cite{zhang2020learning}     & \multirow{5}{*}{fully}   & \xmark & \xmark & -    & 39.81 & 23.25 & -     & 58.75 & 44.05 & 27.38 & -     \\
EMB \cite{huang2022video}        &    & \cmark & \xmark & 72.50 & 58.33 & 39.25 & 53.09 & 64.13 & 44.81 & 26.07 & 45.59 \\
MGSL-Net \cite{liu2022memory}   &    & \cmark & \xmark & -     & 63.98 & 41.03 & -     & 51.87 & 31.42 & -     \\
EaTR \cite{jang2023knowing}       &    & \cmark & \xmark & -     & 68.47 & 44.92 & -     & 58.18 & 37.64 & -     \\
UniVTG \cite{lin2023univtg} &    & \cmark & \xmark  & 72.63 &  60.19 &  38.55 &  52.17 &  & & -     \\
\midrule
CRM \cite{huang2021cross}        & \multirow{4}{*}{weakly}  & \xmark & \xmark & 53.66 & 34.76 & 16.37 & -     & 55.26 & 32.19 & -     & -     \\
CNM \cite{zheng2022weakly_aaai}        &   & \xmark & \xmark & 60.39 & 35.43 & 15.45 & -     & 55.68 & 33.31 & -     & -     \\
CPL \cite{zheng2022weakly_cvpr}        &   & \xmark & \xmark & 66.40 & 49.24 & 22.39 & -     & 55.73 & 31.37 & -     & -     \\
Huang et al. \cite{huang2023weakly} &  & \xmark & \xmark & 69.16 & 52.18 & 23.94 & 45.20 & 58.07 & 36.91 & -     & 41.02 \\
\midrule
Gao et al. \cite{gao2021learning}  & \multirow{5}{*}{unsup.}  & \cmark & \xmark & 46.69 & 20.14 & 8.27  & -     & 46.15 & 26.38 & 11.64 & -     \\
PSVL \cite{nam2021zero}       &   & \cmark & \xmark & 46.47 & 31.29 & 14.17 & 31.24 & 44.74 & 30.06 & 14.74 & 29.62 \\
PZVMR \cite{wang2022prompt}      &   & \cmark & \cmark & 46.83 & 33.21 & 19.14 & 36.15 & 45.63 & 32.14 & 18.71 & 30.35 \\
Kim et al. \cite{kim2023language} &   & \cmark & \cmark & 52.95 & 37.24 & 19.33 & 36.05 & 47.61 & 32.59 & 15.42 & 32.45 \\
SPL \cite{zheng2023generating}        &   & \cmark & \cmark & 60.73 & 40.70 & 19.62 & 40.47 & 50.24 & 27.24 & 15.03 & 35.44 \\
\midrule
GroundingGPT \cite{li2024groundinggpt} & \multirow{3}{*}{fully} & \cmark & \cmark & -     & 29.6  & 11.9  & -     & -     & -     & -     & -     \\
TimeChat-7B \cite{ren2024timechat} &  & \cmark & \cmark & 40.6  & 23.8  & 9.7  & 26.2  & 25.0  & 13.2  & 6.1  & 18.5  \\
VTimeLLM-13B \cite{huang2024vtimellm} &  & \cmark & \cmark & 55.3  & 34.3  & 14.7  & 34.6  & 44.8  & 29.5  & 14.2  & 31.4  \\
\midrule
VideoChat-7B \cite{li2023videochat}  & \multirow{7}{*}{zero-shot} & \cmark & \cmark & 9.0  & 3.3  & 1.3  & 6.5  & 8.8  & 3.7  & 1.5  & 7.2   \\
VideoLLaMA-7B \cite{zhang2023video} &  & \cmark & \cmark & 10.4 & 3.8  & 0.9  & 7.1  & 6.9  & 2.1  & 1.1  & 6.3   \\
VideoChatGPT-7B \cite{maaz2023video} &  & \cmark & \cmark & 20.0 & 7.7  & 1.7  & 13.7 & 26.4 & 13.6 & 6.1  & 18.9  \\
UniVTG \cite{lin2023univtg} &  & \cmark & \xmark & 44.09 & 25.22 & 10.03 & 27.12 & - & - & - & - \\
Luo et al.\cite{wacv} &  & \cmark & \xmark & 56.77 & 42.93 & 20.13 & 37.92 & 48.28 & 27.90 & 11.57 & 30.45 \\
VTG-GPT \cite{xu2024vtg}    &  & \cmark & \cmark & 59.48 & 43.68 & 25.94 & 39.81 & 47.13 & 28.25 & 12.84 & 30.49 \\
TFVTG \cite{tfvtg}  & & \cmark & \cmark & 67.04 & \textbf{49.97} & 24.32 & 44.51 & 49.34 & 27.02 & 13.39 & 34.10 \\
\midrule
\midrule
Ours & zero-shot & \cmark & \xmark & \textbf{67.82} & 48.58 & \textbf{26.67} & \textbf{45.69} & \textbf{51.88} & \textbf{28.91} & \textbf{15.07} & \textbf{36.55}
\\
\bottomrule
\end{tabular}
\caption{Evaluation results on the Charades-STA and ActivityNet Captions datasets.}
\label{table:main_result}
\end{table*}

\begin{table}[t]
\centering
\footnotesize
\renewcommand{\arraystretch}{0.95} 
\begin{tabular}{c|c|ccc}
\toprule
\multirow{2}{*}{Method} & \multirow{2}{*}{Setting} & \multicolumn{3}{c}{Charades-CD} \\ \cline{3-5} \rule{0pt}{1.2EM}
& & R@0.3 & R@0.5 & R@0.7 \\
\midrule
2D-TAN \cite{zhang2020learning}              & fully                   & 43.45 & 30.77 & 11.75 \\
TSP-PRL \cite{wu2020tree}            &   fully  & 31.93 & 19.37 & 6.20  \\
SCDM \cite{yuan2019semantic}               &  fully  & 52.38 & 41.60 & 22.22  \\
WSSL \cite{duan2018weakly}                & weakly                  & 35.86 & 23.67 & 8.27 \\
SPL \cite{zheng2023generating}                 & unsup.                 & 62.96 & 38.25 & 15.53  \\
TFVTG \cite{tfvtg} & zero-shot & 65.07 & 49.24 & 23.05  \\
\midrule
\midrule
Ours & zero-shot & \textbf{67.94} & \textbf{50.19} & \textbf{27.14} \\
\bottomrule
\end{tabular}
\vspace{-0.1cm}
\caption{Results under OOD setting with altered target moment distributions on the Charades-CD dataset}
\label{tab_2}
\end{table}

\section{Experiments Setup}
\subsection{Datasets}
\paragraph{General Settings.}
To verify the effectiveness of our method, we conduct experiments on Charades-STA~\cite{gao2017tall} and ActivityNet Captions~\cite{krishna2017dense} benchmark datasets. 
The Charades-STA dataset is an extension of the original Charades dataset, designed for video-query tasks. It contains 12,408 and 3,720 video-query pairs for the training and test splits, respectively, and we evaluate performance using the test split.
The ActivityNet Captions dataset, initially created for video captioning tasks, consists of 20,000 videos. It includes 37,417, 17,505, and 17,031 video-query pairs in the train, valid-1, and valid-2 splits, respectively. Following previous works~\cite{wacv, wang2022negative, tfvtg}, we evaluate performance using the valid-2 split.

\paragraph{OOD Settings.} 
To demonstrate that our training-free approach effectively enhances robustness and generalization, we conduct experiments across three OOD scenarios: altered target moment distributions, inserted noise moments, and unseen natural language queries.

In the first scenario, we examine the impact of altered target moment distributions. The Charades-STA dataset’s training, validation, and test sets are restructured so that the target moment distribution in the test set differs from that in the training set.

The second scenario examines the effect of inserting noise into the video. We insert randomly generated video segments at the beginning of the test videos to create two types of noise-augmented data, as following DCM~\cite{yang2021deconfounded}. The temporal length of each video is extended to \( \tau + \rho \), and the target moment's timestamps are adjusted to \( (\tau_s + \rho, \tau_e + \rho) \) accordingly. For evaluation, \( \rho \) is set to \( \{10, 15\} \) for Charades-STA and \( \{30, 60\} \) for ActivityNet Captions.
This allows us to evaluate the model's ability to locate target segments based on video context rather than relying on positional patterns.

\begin{table*}[t]
\centering
\footnotesize
\setlength{\tabcolsep}{0.45em} 
\renewcommand{\arraystretch}{0.95} 
\begin{tabular}{c|c|ccc|ccc|ccc|ccc}
\toprule
\multirow{3}{*}{Method} & \multirow{3}{*}{Setting} & \multicolumn{6}{c|}{Charades-STA} & \multicolumn{6}{c}{ActivityNet-Captions} \\ \cline{3-14} \rule{0pt}{1.2EM}
                        &                          & \multicolumn{3}{c|}{$\rho$ = 30} & \multicolumn{3}{c|}{$\rho$ = 60} & \multicolumn{3}{c|}{$\rho$ = 10} & \multicolumn{3}{c}{$\rho$ = 15} \\ 
                        &                          & R@0.5 & R@0.7 & mIoU & R@0.5 & R@0.7 & mIoU & R@0.5 & R@0.7 & mIoU & R@0.5 & R@0.7 & mIoU \\
\midrule
LGI \cite{mun2020local}              & \multirow{5}{*}{fully}                    & 42.1  & 18.6  & 41.2  & 35.8  & 15.4  & 37.1  & 16.3  & 6.2   & 22.2  & 11.0  & 3.9   & 17.3  \\
2D-TAN \cite{zhang2020learning}             &                     & 27.1  & 13.1  & 25.7  & 21.1  & 8.8   & 22.5  & 16.4  & 6.6   & 23.2  & 11.5  & 3.9   & 19.4  \\
MMN \cite{wang2022negative}               &                     & 31.6  & 13.4  & 33.4  & 27.0  & 9.3   & 30.3  & 23.0  & 7.1   & 26.2  & 14.3  & 5.2   & 20.6  \\
VDI \cite{luo2023towards}               &                     & 25.9  & 11.9  & 26.7  & 20.8  & 8.7   & 22.0  & 20.9  & 7.1   & 27.6  & 14.3  & 5.2   & 23.7  \\
DCM \cite{yang2021deconfounded}               &                     & 44.4  & 19.7  & 42.3  & 38.5  & 15.4  & 39.0  & 18.2  & 7.9   & 24.4  & 12.9  & 4.8   & 20.7  \\
\midrule
CNM \cite{zheng2022weakly_aaai}               & \multirow{2}{*}{weakly}                   & 9.9   & 1.7   & 21.6  & 6.1   & 0.5   & 16.6  & 6.1   & 0.4   & 21.0  & 2.5   & 0.1   & 16.8  \\
CPL \cite{zheng2022weakly_cvpr}                &                    & 29.9  & 8.5   & 32.2  & 24.9  & 6.3   & 30.5  & 4.7   & 0.4   & 21.1  & 2.1   & 0.2   & 17.7  \\
\midrule
PSVL \cite{nam2021zero}               & \multirow{2}{*}{unsup.}                  & 3.0   & 0.7   & 8.2   & 2.2     & 0.4     & 6.8     & -     & -     & -     & -     & -     & -     \\
PZVMR \cite{wang2022prompt}              &                   & -     & 8.6     & 25.1     & -   & 6.5   & 28.5  & -     & 4.4     & 28.3     & -   & 2.6   & 19.1  \\
\midrule
Luo et al. \cite{wacv}         & \multirow{2}{*}{zero-shot }               & 40.3  & 18.2  & 38.2  & 38.9  & 17.0  & 37.8  & 18.4  & 6.8   & 21.1  & 18.6  & 7.4   & 20.6  \\
TFVTG \cite{tfvtg} & & \textbf{45.9} & 20.8 & 43.0 & 43.8 & 20.0 & 42.6 & 20.4 & 11.2 & 31.7 & 18.5 & 10.0 & 30.3 \\
\midrule
\midrule
Ours & zero-shot & 45.3 & \textbf{23.2} & \textbf{44.7} & \textbf{44.1} & \textbf{22.0} & \textbf{44.6} & \textbf{28.5} & \textbf{14.7} & \textbf{36.2} & \textbf{28.3} & \textbf{14.5} & \textbf{36.1}\\
\bottomrule
\end{tabular}
\vspace{-0.1cm}
\caption{Results under OOD settings with inserted noise moments on benchmark datasets.}
\label{tab_3}
\end{table*}

\begin{table*}[t]
\centering
\footnotesize
\renewcommand{\arraystretch}{0.95} 
\begin{tabular}{c|c|ccc|ccc}
\toprule
\multirow{3}{*}{Method} & \multirow{3}{*}{Setting} & \multicolumn{6}{c}{Charades-CG} \\ \cline{3-8} \rule{0pt}{1.2EM}
& & \multicolumn{3}{c|}{Insertion Composition Words} & \multicolumn{3}{c}{Insertion Unseen Words}  \\
& & R@0.5 & R@0.7 & mIoU & R@0.5 & R@0.7 & mIoU\\
\midrule
2D-TAN \cite{zhang2020learning}              & \multirow{5}{*}{fully}                & 30.91 & 12.23 & 29.75 &29.36 & 13.21 & 28.47 \\
TSP-PRL \cite{wu2020tree}            &                   & 16.30 & 2.04  & 13.52 &14.83 & 2.61  & 14.03 \\
SCDM \cite{yuan2019semantic}               &                   & 27.73 & 12.25 & 30.84 & -     & -     & -     \\
VISA \cite{li2022compositional}               &                 & 45.41 & 22.71 & 42.03 &42.35 & 20.88 & 40.18 \\
DeCo \cite{yang2023deco}               &                 & 47.39 & 21.06 & 40.70 & -     & -     & -     \\
\midrule
WSSL \cite{duan2018weakly}                & \multirow{2}{*}{weakly}                  & 3.61  & 1.21  & 8.26  & 2.79  & 0.73  & 7.92  \\
CPL \cite{zheng2022weakly_cvpr}                 &                 & 39.11 & 15.60 & 35.53 & 45.90 & 22.88 & -     \\
\midrule
Luo et al. \cite{wacv}         & \multirow{2}{*}{zero-shot}             & 40.27 & 16.27 & -    & 45.04 & 21.44 & -     \\
TFVTG \cite{tfvtg} & & 43.84 & 18.68 & 40.19 & \textbf{56.26} & 28.49 & 46.90 \\
\midrule
\midrule
Ours & zero-shot & 43.55 & \textbf{21.30} & \textbf{41.95} & 52.37 & \textbf{32.66} & \textbf{47.86} \\

\bottomrule
\end{tabular}
\vspace{-0.1cm}
\caption{Results under OOD settings with unseen text queries on the Charades-CG dataset.}
\label{tab_4}
\end{table*}

For the third scenario, we generate unseen words in natural language queries during training with two types of textual noise: Insertion Composition Words, where queries combine words seen during training in novel ways, and Insertion Unseen Words, where queries include new words not encountered during training.
This setup evaluates the model’s ability to generalize to diverse textual queries without over-relying on the training vocabulary.


\section{Experimental Results}
\subsection{Results on General Settings}
Our baselines consist of general video understanding models including VideoChat, VideoLLaMa, VideoChatGPT, and UniVTG~\cite{li2023videochat, zhang2023video, maaz2023video, lin2023univtg}, along with zero-shot VTG models including Luo et al., VTG-GPT, and TGVTG \cite{wacv, vtggpt, tfvtg}.
Table \ref{table:main_result} shows the comparison between our proposed method and recent state-of-the-art VTG methods on Charades-STA and ActivityNet Captions datasets, respectively.
Our proposed method, which does not rely on LLMs, achieves the best performance across all evaluation metrics except for R@0.5 on Charades-STA.
Notably, our method significantly surpasses others with mIoU performance improvements of 2.65\% and 7.18\%.
These results clearly indicate that our method is highly effective in the zero-shot video temporal grounding task.

\begin{figure*}[t]
\centering
\includegraphics[width=0.95\linewidth]{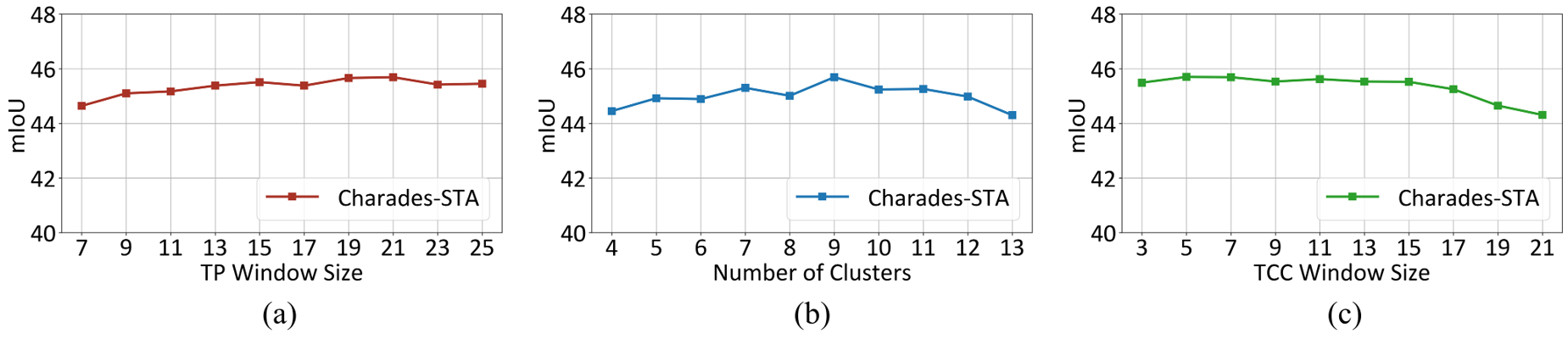}
\vspace{-0.2cm}
\caption{Analysis on hyperparameters. mIoU with respect to (a) temporal pooling window size $w$, (b) number of clusters $k$, and (c) window size $r$ for temporal coherence clustering.}
\label{hyperparameter}
\end{figure*}

\begin{table*}[t!]
\centering
\scriptsize
\footnotesize
\setlength{\tabcolsep}{0.5em} 
\renewcommand{\arraystretch}{1.0} 
\begin{tabular}{c|ccc|ccc|ccc|ccc}
\toprule
\multirow{2}{*}{Setting}
                         & \multicolumn{3}{c|}{Top 25\% Skewness} & \multicolumn{3}{c|}{25-50\% Skewness} & \multicolumn{3}{c|}{50-75\% Skewness} & \multicolumn{3}{c}{75-100\% Skewness} \\ 
                         & R@.5 & R@.7 & mIoU & R@.5 & R@.7 & mIoU & R@.5 & R@.7 & mIoU & R@.5 & R@.7 & mIoU \\

\midrule
    Ours w/o SA & 26.40 & 16.44 & 36.89 & 27.01 & 14.91 & 36.57 & 26.14 & 13.88 & 36.17 & 25.69 & 13.18 & 36.15\\
    Ours & 29.74 & 17.27 & 38.09 & 29.61 & 15.64 & 36.76 & 28.42 & 13.74 & 35.82 & 27.88 & 13.60 & 35.54\\
\midrule
    Difference & +3.34 & +0.82 & +1.21 & +2.61 & +0.73 & +0.20 & +2.28 & -0.14 & -0.34 & +2.18 & +0.42 & -0.61\\

\bottomrule
\end{tabular}
\vspace{-0.1cm}
\caption{Analysis of the effect of similarity adjustment on partitions ordered by similarity score skewness, from highest to lowest.}
\label{anal_3}
\end{table*}

\subsection{Results on OOD settings} 
Our proposed method demonstrates superior performance across all metrics in generalization, as shown in \cref{tab_2}. Since it effectively captures the temporal context of the given video, it can generate segment proposals that adapt to the video's temporal dynamics. Consequently, our method can generate accurate segments regardless of the statistical priors of the actual video segment data.

\Cref{tab_3} presents the experimental results for the noise insertion video OOD setting. Overall, zero-shot approaches demonstrate stronger robustness compared to training-based methods under noise-inserted scenarios.
For Charades-STA dataset, our proposed method demonstrates the best performance across all evaluation metrics except for R@0.5 with \( \rho = 30 \).
In contrast, for the ActivityNet-Captions dataset, our method significantly outperforms all metrics. The mIoU improvements are 14.20\% and 19.14\%, respectively.
This demonstrates that our method achieves significant robustness against noise-inserted scenarios.
It shows that our proposed method effectively aggregates consecutive features, which helps to mitigate the influence of noise-inserted segments.

In \cref{tab_4}, we observed that our proposed method shows better mIoU performance under two types of textual noise scenarios. 
For the insertion composition words scenario, our method consistently demonstrates good performance, and for the insertion unseen words scenario, it performs well in most cases.
This result highlights that simply employing LLMs for query diversification is insufficient; instead, incorporating robust feature extraction and temporal context is crucial for handling noisy queries effectively.

\begin{table}[t]
\centering
\footnotesize
\setlength{\tabcolsep}{0.2em}          
\renewcommand{\arraystretch}{1}       
\begin{tabular}{ccc|cccc|cccc}
\toprule \rule{0pt}{0.0EM}
\multirow{2}{*}{TP} & \multirow{2}{*}{TCC} & \multirow{2}{*}{SA}  & \multicolumn{4}{c|}{Charades-STA} & \multicolumn{4}{c}{ActivityNet-Captions} \\ \cline{4-11} \rule{0pt}{1.2EM}
 & & & R@0.3 & R@0.5 & R@0.7 & mIoU & R@0.3 & R@0.5 & R@0.7 & mIoU
\\ \midrule \rule{0pt}{0.0EM}
\xmark & \xmark & \xmark & 60.99 & 40.83 & 19.06 & 39.71 & 49.78 & 26.63 & 14.83 & 35.49
\\ \rule{0pt}{0.0EM}
\cmark & \xmark & \xmark & 65.16 & 47.34 & 24.38 & 43.91 & 50.90 & 26.32 & 14.72 & 36.41
\\ \rule{0pt}{0.0EM}
\cmark & \cmark & \xmark & 67.74 & 48.01 & 26.34 & 45.44 & 50.81 & 26.31 & 14.60 & 36.44
\\ \midrule \rule{0pt}{0.0EM}
\cmark & \cmark & \cmark & \textbf{67.82} & \textbf{48.58} & \textbf{26.67} & \textbf{45.69} & \textbf{51.88} & \textbf{28.91} & \textbf{15.07} & \textbf{36.55}\\ \bottomrule
\end{tabular}
\vspace{-0.1cm}
\caption{Performance with our proposed modules. TP, TCC, and SA represent the temporal pooling, temporal coherent clustering, and similarity adjustment, respectively.}
\label{ablation}
\end{table}

\subsection{Ablation Study}
\Cref{ablation} presents the performance of our method on both datasets with different combinations of our proposed modules.
This demonstrates that our proposed method, which incorporates temporal pooling, temporal coherence clustering, and similarity adjustment, effectively generates high-quality proposals and localizes proposals more accurately.

\subsection{Further Analysis}
\paragraph{Hyperparameters.}
To validate the sensitivity of the hyperparameters, we conducted experiments on Charades-STA by varying the window size \(w\) for temporal pooling, the number of clusters \(k\) for temporal coherence clustering, and the window size \(r\) for temporal coherence clustering.
The results are shown in \cref{hyperparameter}.
Overall, these results demonstrate that our method is not sensitive for all hyperparameters. For the Charades dataset, the optimal values were set to \(w = 21\), \(k = 9\), and \(r = 7\). Importantly, we did not tune hyperparameters separately for each dataset; instead, we used the same values for all experiments across both Charades-STA and ActivityNet.

\begin{table}[t]
\centering
\footnotesize
\setlength{\tabcolsep}{0.5em}          
\renewcommand{\arraystretch}{1.0}
\begin{tabular}{c|cccc}
\toprule
Feature Extraction & R@0.3 & R@0.5 & R@0.7 & mIoU \\
\midrule
Gaussian ($\sigma=1$) & 65.27 & 47.72 & 24.27 & 43.96 \\ 
Gaussian ($\sigma=4$) & 67.63 & 48.44 & 26.24 & 45.36 \\ \midrule
TP (Ours) & \textbf{67.82} & \textbf{48.58} & \textbf{26.67} & \textbf{45.39} \\
\bottomrule
\end{tabular}
\vspace{-0.1cm}
\caption{Analysis of temporal pooling (TP)}
\label{anal_1}
\end{table}

\begin{table}[t]
\centering
\footnotesize
\setlength{\tabcolsep}{0.5em}          
\renewcommand{\arraystretch}{1.0}
\begin{tabular}{c|cccc}
\toprule
Clustering & R@0.3 & R@0.5 & R@0.7 & mIoU \\ \midrule 
K-Means & 65.40 & 47.74 & 24.92 & 44.22\\
TW-FINCH & 66.83 & 47.34 & 24.78 & 44.67\\ \midrule 
TCC (Ours) & \textbf{67.82} & \textbf{48.58} & \textbf{26.67} & \textbf{45.69} \\ \bottomrule
\end{tabular}
\vspace{-0.1cm}
\caption{Analysis of temporal coherence clustering (TCC) }
\label{anal_2}
\end{table}

\paragraph{Temporal pooling.}
While we uniformly pooled frame features over adjacent frames, other methods, such as gaussian pooling, can also be applied. 
\Cref{anal_1} presents the results with a gaussian approach, showing that as the sigma value increases, the results become more similar to our method.

\paragraph{Temporal coherence clustering.}
As shown in \cref{anal_2}, our proposed temporal coherence clustering achieves higher performance compared to other clustering methods~\cite{hartigan1979algorithm, twfinch}. This demonstrates that enforcing temporal consistency during clustering leads to more coherent segment boundaries, thereby improving proposal quality and localization accuracy.

\paragraph{Similarity adjustment.}
Since the ActivityNet dataset consists of diverse scenes, it exhibits higher similarity distribution skewness. Therefore, we analyze the effect of similarity adjustment across different levels of skewness on ActivityNet, as shown in \cref{anal_3}. Our method shows strong improvements under highly skewed conditions. This confirms that adjusting similarity distributions is crucial for robust proposal scoring and optimal segment selection.

\section{Conclusion}
In this paper, we presented TAG, a simple yet effective Temporal-Aware approach for zero-shot video temporal grounding.
By incorporating temporal pooling, temporal coherence clustering, and similarity adjustment, our proposed method effectively captured the temporal context of videos and enhanced temporal coherence without additional training. 
Our approach achieved state-of-the-art results on the Charades-STA and ActivityNet benchmark datasets, without reliance on LLMs.
{
    \small
    \bibliographystyle{ieeenat_fullname}
    \bibliography{main}
}

\clearpage
\setcounter{page}{1}
\maketitlesupplementary
\setcounter{section}{0}
\renewcommand{\thesection}{\Alph{section}}


\section{Implementation Details}
All experiments are conducted using the PyTorch framework on an NVIDIA RTX 3090Ti GPU. To ensure a fair comparison, all conditions are set according to the protocols outlined in existing zero-shot VTG methods~\cite{wacv, vtggpt, tfvtg}. 
To extract image and text features, we adopt the pretrained BLIP-2 Q-former~\cite{li2023blip} as the VLM.
For both datasets, we set window size \(w\) to 21 for temporal pooling, the number of clusters to 9 for temporal coherence clustering, and the window size \(r\) to 7 for temporal coherence clustering. Notably, we used the same values for all experiments across both Charades-STA and ActivityNet.
To adjust similarity adjustment, we apply Box-Cox transformation~\cite{box1964analysis} with \( \lambda\), which is automatically determined from the scikit-learn tool~\cite{schikit}. 

\section{Evaluation Metrics}
\label{metric}
We adopt the evaluation metrics \textit{R@m} and \textit{mIoU} in the previous work~\cite{wacv, wang2022negative, tfvtg}. Here, \textit{m} refers to a predefined temporal Intersection over Union (IoU) threshold. Specifically, \textit{R@m} denotes the proportion of predicted moments with IoU values exceeding the threshold m, while \textit{mIoU} represents the average Intersection over Union across all predictions. The evaluation metrics are formally defined as follows:
\begin{equation}
    \text{R@m} = \frac{1}{N} \sum_{i=1}^{N} \mathds{1}(\text{IoU}(P_i, G_i) > m)
\end{equation}
\begin{equation}
    \text{IoU}(P_i, G_i) = \frac{\text{Intersection}(P_i, G_i)}{\text{Union}(P_i, G_i)}
\end{equation}
\begin{equation}
    \text{mIoU} = \frac{1}{N} \sum_{i=1}^{N} \text{IoU}(P_i, G_i)
\end{equation}
where $N$ is the total number of predicted moments, $P_i$ is the predicted temporal segment for the $i$-th sample, $G_i$ is the ground truth temporal segment for the $i$-th sample, $m$ is the predefined IoU threshold, and $\mathds{1}(\cdot)$ is the indicator function that equals 1 if the condition is true and 0 otherwise.

\section{Ablations on Similarity Adjustment}
\Cref{tab_12} presents the results of replacing only the normalization strategy. In fact, our method does not specifically rely on the Box-Cox transformation. Adjusting a skewed similarity distribution consistently improves performance, regardless of the normalization strategy.

\begin{table}[h]
\centering
\scriptsize
\setlength{\tabcolsep}{0.3em}          
\renewcommand{\arraystretch}{1.0}       
\begin{tabular}{c|cccc|cccc}
\toprule \rule{0pt}{0.0EM}
Normalization & \multicolumn{4}{c|}{Charades-STA} & \multicolumn{4}{c}{ActivityNet-Captions} 
\\ \cline{2-9} \rule{0pt}{1.2EM}
Method & R@0.3 & R@0.5 & R@0.7 & mIoU & R@0.3 & R@0.5 & R@0.7 & mIoU
\\ \midrule \rule{0pt}{0.0EM}
None & 67.74 & 48.01 & 26.34 & 45.44 & 50.81 & 26.31 & 14.60 & 36.44 \\
Yeo-Johnson & 67.42 & 47.74 & 25.97 & 45.27 & 51.29 & 28.28 & 14.96 & 36.39 \\ \midrule
Box-cox (Ours)  & \textbf{67.82} & \textbf{48.58} & \textbf{26.67} & \textbf{46.69} & \textbf{51.88} & \textbf{28.91} & \textbf{15.07} & \textbf{36.55}\\ \bottomrule
\end{tabular}
\vspace{-0.1cm}
\caption{Performance with normalization methods}
\label{tab_12}
\end{table}

\section{Ablations on the VLMs}
To extract alignment similarities, \citet{wacv} uses InternVideo, while VTG-GPT and TFVTG \cite{vtggpt, tfvtg} use BLIP-2 as VLMs.
To ensure a fair comparison, we evaluate the performance of different VLMs, as presented in \cref{appendix_c}.
It can be observed that using BLIP-2 as the VLM achieves the best performance across all metrics.
Also, our method outperforms existing approaches that use the same VLM \cite{wacv,vtggpt,tfvtg}.
These results indicate that our method is effective regardless of the VLM used.

\begin{table}[h]
\centering
\aboverulesep=0ex 
\belowrulesep=0ex 
\footnotesize
\renewcommand{\arraystretch}{1.0}       
\begin{tabular}{c|cccc}
\toprule \rule{0pt}{1.0EM}
VLMs & R@0.3 & R@0.5 & R@0.7 & mIoU \\
\midrule \rule{0pt}{1.0EM}
CLIP & 60.97 & 42.80 & 22.47 & 40.51 \\
\rule{0pt}{1.0EM}
InternVideo & 62.34 & 43.28 & 22.20 & 41.48 \\
\midrule \rule{0pt}{1.0EM}
BLIP-2 (Ours) & \textbf{67.82} & \textbf{48.58} & \textbf{26.67} & \textbf{46.69}  \\
\bottomrule
\end{tabular}
\vspace{-0.1cm}
\caption{Evaluation results of ours on Charades-STA with different VLMs}
\label{appendix_c}
\vspace{-0.5cm}
\end{table}

\section{About the Use of LLMs}
Our proposed method achieves superior performance without relying on LLMs. When incorporating LLMs, its performance is further improved, as shown in \cref{LLM}.
\Cref{prompt} presents the prompts that we use. We utilize GPT-4 to generate paraphrased versions of the given queries. Using these augmented queries, we extract temporally aggregated features and generate temporal-aware proposals.
Subsequently, among the proposals generated for the original query and its augmented versions, we select the final proposal based on equation (6).

\begin{table}[h]
\centering
\scriptsize
\setlength{\tabcolsep}{0.3em}          
\renewcommand{\arraystretch}{1.0}       
\begin{tabular}{c|cccc|cccc}
\toprule \rule{0pt}{0.0EM}
\multirow{2}{*}{Method}  & \multicolumn{4}{c|}{Charades-STA} & \multicolumn{4}{c}{ActivityNet-Captions} 
\\ \cline{2-9} \rule{0pt}{1.2EM}
 & R@0.3 & R@0.5 & R@0.7 & mIoU & R@0.3 & R@0.5 & R@0.7 & mIoU
\\ \midrule \rule{0pt}{0.0EM}
Ours & 67.82 & 48.58 & 26.67 & 45.69 & 51.88 & 28.91 & 15.07 & 36.55 \\
Ours w/ LLM & 69.35 & 49.95 & 26.59 & 46.42 & 51.89 & 28.88 & 15.12 & 36.61 \\ \bottomrule
\end{tabular}
\vspace{-0.1cm}
\caption{Performance analysis on the effect of incorporating LLMs into our method}
\label{LLM}
\end{table}

\begin{table*}[h]
\centering
\aboverulesep=0ex 
\belowrulesep=0ex 
\footnotesize
\setlength{\tabcolsep}{0.45em}     
\begin{tabular}{|p{\textwidth}|}
\toprule \rule{0pt}{1.0EM}
Instruction: \\
Your task is to analyze the user's query. You need to provide multiple textual descriptions for each query as comprehensively as possible. 
You can diversify the sentence structure and word usage, but you should strictly keep the same semantic meaning. \\ 
\\
~Example: \\
1.  \\
- User Input: "a person is sitting in front of a computer sneezing." \\ 
- Generated result:  
  "An individual is seated at a desk, sneezing while using a computer.",  
  "Someone is in front of a computer, sneezing as they sit.", 
  "A person sneezes while sitting in front of their computer." \\
2.  \\
- User Input: "A person sits on a couch."  \\
- Generated result:  
  "A person is sitting on a couch.",  
  "An individual takes a seat on a sofa.",  
  "Someone sits down on a couch."  \\
3.  \\
- User Input: "A person runs around the room in a circle."  \\
- Generated result:  
  "A person is running around the room in a circular pattern.",  
  "An individual is observed jogging in a circle within a room.", 
  "Someone runs in circles around the room." \\
4.  \\
- User Input: "Two people are knelled in front of the man." \\ 
- Generated result:  
  "Two individuals are kneeling in front of a man.", 
  "A pair of people are knelt before a man.", 
  "Two persons kneel in front of a man."  \\
  
\bottomrule
\end{tabular}
\vspace{0.1cm}
\caption{Prompt Examples. When we incorporate the LLM, we generate paraphrased versions of the given queries using the prompt examples.}
\label{prompt}
\vspace{0.1cm}
\end{table*}

\begin{figure*}[h]
\flushleft
\includegraphics[width=0.9\linewidth]{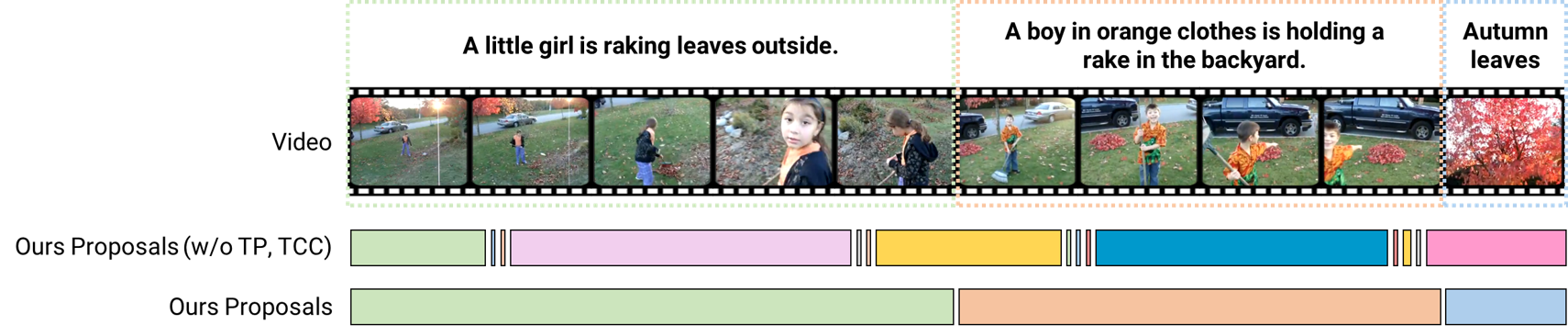}
\caption{Qualitative results on ActivityNet Captions dataset.}
\label{qualitative}
\end{figure*}

\begin{table}[h]
\centering
\footnotesize
\setlength{\tabcolsep}{0.8em}          
\renewcommand{\arraystretch}{1.0}       
\begin{tabular}{cc|c|c}
\toprule \rule{0pt}{0.0EM}
\multirow{2}{*}{TP} & \multirow{2}{*}{TCC}  & Charades-STA & ActivityNet-Captions \\ \cline{3-4} \rule{0pt}{1.2EM}
 & & \# of clusters per GT~($\downarrow$) & \# of clusters per GT~($\downarrow$)
\\ \midrule \rule{0pt}{0.0EM}
\xmark & \xmark & 7.90 & 24.32
\\ \rule{0pt}{0.0EM}
\cmark & \xmark & 3.43 &  5.13
\\ \rule{0pt}{0.0EM}
\xmark & \cmark & 3.54 & 6.74
\\ \rule{0pt}{0.0EM}
\cmark & \cmark & 3.57 & 4.65
\\ \bottomrule
\end{tabular}
\vspace{-0.1cm}
\caption{Analysis of semantic fragmentation}
\label{semantic fragmentation}
\vspace{-0.5cm}
\end{table}

\section{Analysis of Semantic Fragmentation}
When semantic fragmentation is reduced, the number of generated segments within each ground-truth moment is expected to decrease. To analyze the effectiveness of our method in mitigating semantic fragmentation, we measure the number of segments within each ground-truth segment. With our method applied, this measure is significantly reduced, as shown in \cref{semantic fragmentation}. It indicates that our approach effectively captures the video's temporal context and enhances temporal coherence, thereby enabling accurate localization of the target moment.

Furthermore, we present qualitative results, as shown in \cref{qualitative}. It supports the effectiveness of our proposed temporal pooling and temporal coherence clustering.

\section{Qualitative results}
\begin{figure*}[h]
    \centering
    {
    \includegraphics[width=0.75\linewidth]{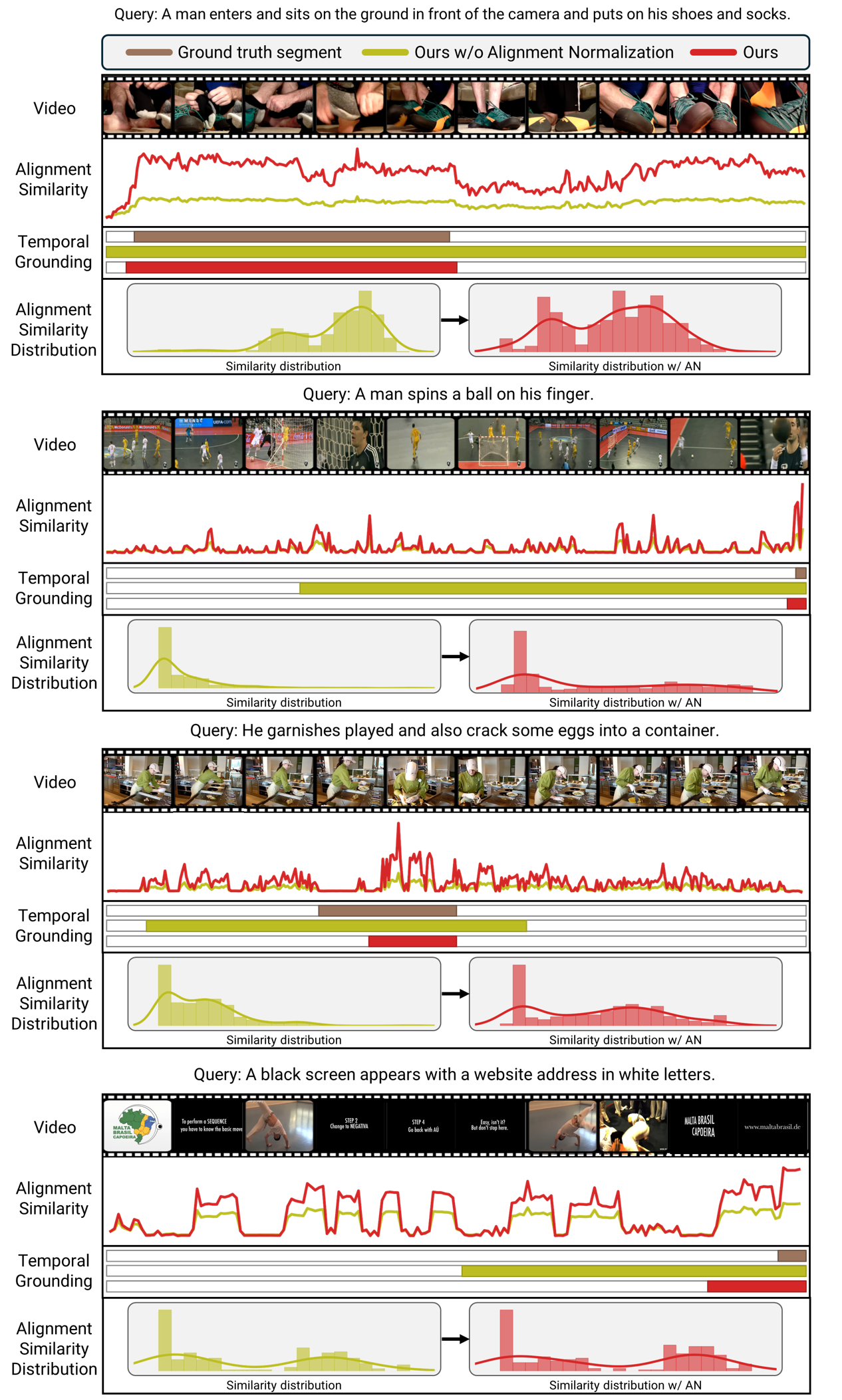}
    \label{fig7}
    }
    \caption{Qualitative results on ActivityNet Captions dataset.}
    \label{more_qualitative}
\end{figure*}

\Cref{more_qualitative} shows the results of our proposed method with and without similarity adjustment against the ground truth moment. When basic similarities are used, the similarity distribution tends to be skewed. The score gap between the target moment and other candidates becomes less significant, making it difficult to accurately identify the optimal segment. In contrast, when similarity adjustment is applied, it enables adaptive normalization based on the similarity distribution, leading to more accurate moment prediction.

\end{document}